\documentclass[3p,twocolumn]{elsarticle}

\let\today\relax
\makeatletter
\def\ps@pprintTitle{%
    \let\@oddhead\@empty
    \let\@evenhead\@empty
    \def\@oddfoot{\footnotesize\itshape
         {Accepted in Pattern Recognition} \hfill\today}%
    \let\@evenfoot\@oddfoot
    }
\makeatother

\usepackage{amsmath,amsfonts}
\usepackage{url}
\usepackage{graphicx}
\graphicspath{{figures/}}
\usepackage{amssymb}
\usepackage{bm}
\usepackage{booktabs}
\usepackage{siunitx}
\newcommand{\comment}[1]{}
\usepackage{xcolor}
\renewcommand{\vec}[1]{\mathbf{#1}}

\newcommand{\method}[0]{TreEnhance}

\begin{document}

\begin{frontmatter}

\title{\method: A Tree Search Method For Low-Light Image Enhancement}

\author[1]{Marco Cotogni\corref{cor1}}
\ead{marco.cotogni01@universitadipavia.it}
\author[1]{Claudio Cusano}
\ead{claudio.cusano@unipv.it}
\cortext[cor1]{Corresponding author}

\address[1]{Dep.\ of Electrical, Computer and Biomedical Engineering, University of Pavia, Via Ferrata 1, Pavia, 27100, Italy}

\begin{abstract}
 In this paper we present \method{}, an automatic method for low-light image enhancement capable of improving the quality of digital images.  The method combines tree search theory, and in particular the Monte Carlo Tree Search (MCTS) algorithm, with deep reinforcement learning. Given as input a low-light image, \method{} produces as output its enhanced version together with the sequence of image editing operations used to obtain it. During the training phase, the method repeatedly alternates two main phases: a generation phase, where a modified version of MCTS explores the space of image editing operations and selects the most promising sequence, and an optimization phase, where the parameters of a neural network, implementing the enhancement policy, are updated. 
 
 Two different inference solutions are proposed for the enhancement of new images: one is based on MCTS and is more accurate but more time and memory consuming; the other directly applies the learned policy and is faster but slightly less precise. As a further contribution, we propose a guided search strategy that ``reverses'' the enhancement procedure that a photo editor applied to a given input image.
 
Unlike other methods from the state of the art, \method{} does not pose any constraint on the image resolution and can be used in a variety of scenarios with minimal tuning. We tested the method on two datasets: the Low-Light dataset and the Adobe Five-K dataset obtaining good results from both a qualitative and a quantitative point of view.

\end{abstract}

\end{frontmatter}

\section{Introduction}

Image enhancement is a well-known image processing problem that consists in improving the appearance of a low quality image to obtain a high quality version without changing the content. Nowadays, Deep Convolutional Neural Networks are the methods of choice for many image processing tasks such as image enhancement, super resolution, image inpainting, etc.  
For this reason, neural networks are now part of many of the image processing tools used by professional photographers to post-process their pictures.  However, the use of this kind of tools requires a certain level of experience and knowledge, making desirable to have both semi-automatic and fully automatic procedures.

Among neural models, image-to-image translation methods
\cite{isola2017image,zhu2017unpaired,ignatov2017dslr} have been shown to be very effective for image enhancement.  However, they suffer of two major problems: the presence of artifacts their output images, and the limitation on the resolution of the input that is often imposed by the architecture of the neural network.
Among these translation approaches, recent methods based on diffusion models seem able to produce high-quality images without artifacts~\cite{saharia2022palette,kim2022diffusionclip,sasaki2021unit,batzolis2021conditional}. These methods repeat many generation steps, preserving every time the spatial resolution of the image. As a consequence, their computational cost can be prohibitive when generating medium- or high-resolution images.

Most recent methods work as black boxes, without providing any insight about the transformation found by the neural network.  For a professional usage, it is of paramount importance that the method provides an ``explainable'' solution that makes it possible for the user to verify it and possibly to adjust it tow his or her needs. Possible solutions to provide an explanation to the neural networks' predictions were explored in the recent years \cite{lundberg2017unified,selvaraju2017grad}. In these works the authors proposed methods to analyze the predictions obtained by the model and obtain some explanation behind the networks' reasoning.


On the basis of these considerations, we propose here \method{}, a low-light image enhancement method based on tree search and deep reinforcement learning.  The method is able to obtain high quality output images with the application of a sequence of image editing operations.
The sequence selected by the method is fully explainable, making it suitable not only for automatic enhancement, but also as a tool for a variety of interactive scenarios.  For instance, it could be used for educational purposes to teach beginners how to edit their own images.

The method does not pose any constraint on the resolution of input images. In fact it can deal with both high- and low-resolution images without any modification. We tested our method on two different datasets tuning only the parameters of the image editing operations.

The main contributions presented in this paper are:
\begin{itemize}
    \item To the best of our knowledge, this is the first work presenting the application of tree-search theory to image enhancement problems.
    \item Differently from several previous methods in the state of the art, \method{} acts as a fully explainable neural architecture for image enhancement. This method provides along with the resulting output image the sequence of enhancing operators selected by the algorithm to enhance the low quality image. The images provided as output are good from a qualitative and quantitative points of view without the introduction of any artifact and without constraints on the image resolution.
    \item \method{} includes two different inference strategies for enhance low quality image based on the data, time an memory constraints.
    \item A guided search procedure that can be used as a tool for ``reverse engineering''. It decomposes into simple operations the enhancement process selected by an expert photographer. This tool could be included in an educational software to teach beginners how to enhance their own photographs.
\end{itemize}

The paper is organized as follows: we begin with the presentation of the state of the art in image enhancement in Section \ref{sec:related}. In Section \ref{sec:method} we explain how \method{} works, what are the steps of the method and which neural architecture it uses. In Section \ref{sec:exp_res} we describe the datasets used for the experiments and the results obtained. Finally we conclude the paper with Section \ref{sec:discuss} and Section \ref{sec:concl} where we discuss the results obtained and possible future investigations on this topic. 
Code is available here: \textcolor{blue}{https://github.com/OcraM17/TreEnhance}

\section{Related Work}
\label{sec:related}
In recent years, many image enhancement methods have been proposed. Among these, one of the most effective approaches is that represented by image-to-image translation methods.  In this family of methods, a neural network learns how to directly convert an image from one domain to another (i.e.\ low quality to high quality), while preserving its visual content. These methods learn by observing thousands of pairs of low-quality/high-quality images.

Ravirathinam et al. proposed a multi-context framework based on a modified version of the UNet architecture for low-light image enhancement \cite{ravirathinam2021c}. In this work the authors combined a perceptual loss, a structural loss and a patch-wise euclidean loss to enhance a low-light input image.

Xu et al. presented a decomposition and enhancement method for low-light image enhancement working in the frequency domain. The architecture presented learns how to recover image objects low-frequency layers. Once these image objects have been recovered, the method is able to enhance high-frequency image details.

Ignatov et al. presented an image-to-image translation pipeline for enhancing smartphone pictures \cite{ignatov2017dslr}. In this work the ground truth images were obtained with a professional camera. In order to deal with misalignment problems due to the different resolutions of the cameras, the authors trained the neural model with aligned patches obtained from the original images.

Pix2Pix \cite{isola2017image},Cycle-GAN\cite{zhu2017unpaired} and EnlightenGAN \cite{jiang2021enlightengan} are methods based on an adversarial learning schema which involves a generator and a discriminator. In a image-to-image translation scenario, the generator learns how to produce images very similar to the training ones, the discriminator learns how to distinguish real training images from generated images. The optimal point is reached when the discriminator is not able to distinguish the images in the training from those produced by the generator. These works show very good performance in several computer vision tasks including image enhancement.

Lv et al. proposed a method based on local features extractors for low light images \cite{lv2018mbllen}. Lore et al. proposed a deep autoencoder approach for natural low-light image enhancement.\cite{lore2017llnet}

Zhang proposed a pipeline based on decomposition network and illumination adjustment to tune the exposure in low-light images \cite{zhang2019kindling}. The same approach has been further explored by other research groups in the following years \cite{lv2021low,zhu2020zero}.


Cai et al.\ presented a Pixel-level Noise-aware Generative Adversarial Network able to generate very realistic noisy images.  By fine-tuning different denoising methods, they were able to reach state-of-the-art performance. This confirms the ability of adversarial networks in generating a very realistic noise in terms of distribution and intensity~\cite{cai2021learning}.

In the last years, methods based on diffusion models reached state-of-the-art performance in image generation, with the additional feature of limiting the artifacts in the output. Compared with GANs, diffusion models are more stable and the images produced obtained a higher FID~\cite{dhariwal2021diffusion}.
Saharia et al.\ proposed a conditional diffusion model and applied it to four different image-to-image translation tasks: colorization, JPEG restoration, inpainting and uncropping~\cite{saharia2022palette}.
Batzolis et al. proposed a conditional diffusion model able to reach state-of-the-art performance in inpainting, super-resolution and edge-to-image tasks~\cite{batzolis2021conditional}.
Similarly to Cycle-GAN, Sasaki et al. proposed an unpaired approach with denoising diffusion probabilistic models for image-to-image translation~\cite{sasaki2021unit}.

Despite their good performance, the application of diffusion models to the enhancement of low-light images is still unexplored. Moreover, according to Dhariwal et al.~\cite{dhariwal2021diffusion}, the computational resources required to train and to use this kind of methods are very high, due to the many refinement steps in the generation process and the necessity of keeping at each step the desired spatial resolution of the images.

Differently from image-to-image translation methods, parametric methods do not model directly the mapping between low-quality and high-quality images but they learn instead the parameters of the color transformation to be applied to the low-quality image. 

Bianco et al. presented two works: in the first one, a neural network learns the parameters of a color transformation. This transformation is applied to a filtered image to recover its original color curve. In the second one, the parameters of a color transformation are combined with a basis function to enhance the visual content of low quality images \cite{bianco2017artistic,bianco2019learning}. Another work from the same authors explores the use of a neural network to estimate the coefficients of splines, that are the used as color curves~\cite{bianco2020personalized}. 

Zhang et al. proposed an Exposure Correction Network able to estimate the best S-shaped curve (a non linear curve used to adjust the exposure of shadow-tone areas in images) to restore low-light images. This curve is then applied to the low quality input image to enhance it \cite{zhang2019zero}.

Chai et al. presented an approach for parametric color enhancement. In this work, a convolutional neural network learns the parameters of a quadratic color transformation in a supervised learning scenario \cite{chai2020supervised}.

Kim et al. proposed a method based on representative color transformations. This method uses local and global enhancement modules to determine the most representative colors in input images and to estimate the transformation for these colors. Then, the model defines the enhanced colors by using the transformed colors. The criteria used to define the enhanced colors is based on the similarity between representative and input colors \cite{kim2021representative}.

Guo et al. presented a zero-reference method for low-light image enhancement. In this work, a deep curve estimator network takes as input a low-light image and it estimates a series of light-enhancement curves. This curves are then iteratively applied to the RGB input image obtaining the enhanced image. In order to train the estimator in a zero-reference scenario, four differentiable non-reference losses were involved \cite{guo2020zero}.

Another kind of related approaches, are those based on reinforcement learning. In this family of methods an agent selects the enhancement operators to be applied to a low quality image to obtain its enhanced version. During the training a numerical reward is used to improve the agent. 

Park et al. presented a deep q-learning based approach for image enhancement. The reward is modeled as the difference between the distance of the image from the ground truth before and after the application of the operator \cite{park2018distort}.

Yang et al. proposed a method based on Markov Decision Process for real-time exposure control. In this work, given the current frame to the agent, a fully convolutional neural network is trained by using the Gausssian policy gradient algorithm.  The method is able to optimize the trade-off among convergence, minimal temporal oscillation and quality of the produced image \cite{yang2018personalized}.

Hu et al. proposed a white box-approach for image enhancement. This work is based on an actor-critic algorithm to enhance the content of a low quality image \cite{hu2018exposure}. 

Yu et al. presented a deep q-learning based tool-chain for image restoration. In this work, a neural network learns what are the most suitable restoration operators that should be applied to a corrupted image to restore its content. \cite{yu2018crafting}.


Yu et al. proposed a mixed approach based on GANs and deep reinforcement learning \cite{yu2018deepexposure}.

The use of vision transformers for color transition problems has been proposed by Cai et al.~\cite{cai2022mask} and by Lin et al.~\cite{lin2022coarse}.  The former proposed a spectral-wise Multi-head Self-Attention block for hyperspectral image reconstruction within a masked transformer. The latter presents a transformer-based method that uses a coarse patch selection in combination with a fine pixel clustering for the reconstruction of hyperspectral images.

Finally, a new approach presented by Zhang et al.\ is based on a transformer architecture for enhancing low quality images. This algorithm, divides the images in patch, applies an embedding and passes the created vectors through several two-branches transformer modules obtaining the enhanced image~\cite{zhang2021star}.

\section{Method}
\label{sec:method}

\method{} is a tree-based deep reinforcement learning method that finds sequences of image editing operations to improve the quality of digital photographs.  It does so by emulating the behavior of a professional photo editor who selects the operations among those available in the software he or she uses.  The selection follows a trial-and-error strategy in which multiple operations (such as gamma correction, contrast adjustment, etc.) are repeatedly tried, verified, canceled, and combined.  This way many different sequences of operations are evaluated and the most promising one is taken.   

More precisely, given an image $\vec{x}$ the method selects the editing operations $a_1, \dots, a_n$ from a finite set $\mathcal{A}$ of operations, and apply them to produce the enhanced image $\vec{\hat{y}}$:
\begin{equation}
    \hat{\vec{y}} = a_n(a_{n - 1}(\dots a_1(\vec{x}))).
\end{equation}
The sequence is terminated by a special \texttt{STOP} operation.

To select the operations, the method builds a search tree that explores a large number of candidate sequences.
A convolutional neural network guides the growth of the tree by promoting the exploration of the operations that are more likely to contribute to the improvement of the image.  The network also evaluates the images obtained during the process so that the best one can be selected.

The network is trained in a supervised way with     the goal of obtaining an enhanced image $\hat{\vec{y}}$ as close as possible to a target image $\vec{y}$ manually edited by a human expert. 
The following sections present the method in details.

\

\subsection{Training Procedure}
\label{sec:MCTS}
A set of input/target image pairs is used to train the neural network.
The training procedure alternates two phases: generation and optimization.
In the generation phase a revised version of the Monte Carlo Tree Search (MCTS) algorithm is used to grow trees associated to a random selection of the available images pairs.
In the optimization phase the images obtained in the generation phase are used to train the parameters of the neural network.

\subsubsection{Generation}
starting from the root node, which represents the input image, the method selects an image editing operation (an action) which results in a new image (i.e. a new node). Nodes and actions form a tree which is built by a sequence of iterations, each one adding a new node.  A numerical score, called \emph{return} is associated to each node. The goal is to find actions that are likely to lead to nodes associated to high returns.
Each iteration of MCTS includes four major steps~\cite{browne2012survey}:
\begin{enumerate}
    \item \textbf{Selection}: a \textit{tree policy} is applied to select a leaf node. 
    \item \textbf{Expansion}: new child nodes (one for each possible action) are attached to the selected leaf.  They will all become new leaves of the tree.
    \item \textbf{Evaluation}: the return is assigned to the node.
    \item \textbf{Backup}: the return is propagated from the selected node up to the root of the tree.
\end{enumerate}
In the original MCTS the evaluation step consists in the repeated application of a random rollout policy until a terminal node is found and the corresponding return is obtained. 

Here, similarly to the method proposed by Silver et al.~\cite{silver2018masteringzero}, the nodes chosen in the selection step are processed by a convolutional neural network.
The network takes as input an image $\vec{x}$ and computes a policy $\pi(\vec{x}, \cdot)$ and a value $v(\vec{x})$.
The policy estimates the probability distribution of the best operation over all the available editing operations ($\pi(\vec{x}, a)$ is the probability estimate that $a$ is the best operation for $\vec{x}$).
The value, instead, estimates the expected return $r(\cdot)$ obtained by the descendants of $\vec{x}$.

The tree policy used in the selection step consists in the maximization of the Upper Confidence Bound (UCB) which accounts for both the exploration of the less frequently visited nodes and the exploitation of those that obtained high returns on average:  
\begin{equation}
    \label{eq:ucb}
    \operatorname{UCB}(\vec{x}) = \bar{r}(\vec{x}) + c \cdot \pi(\vec{x}_p, a) \frac{\sqrt{N(\vec{x}_p)}}{N(\vec{x}) + 1},
\end{equation}
where $\bar{r}(\vec{x})$ is the average return that has been propagated through $\vec{x}$ in previous iterations, $a \in \mathcal{A}$ is the operation that produced $\vec{x}$ from its parent $\vec{x}_p$ (i.e.~$\vec{x} = a(\vec{x}_p$)), and $N(\cdot)$ denotes the number of times a given node has been visited in previous selection steps (that is, $N(\vec{x}_p)$ and $N(\vec x)$ are the number of visits of the parent and the child node, respectively).
The coefficient $c$ balances the exploitation and the exploration terms.

The selection step starts from the root and descends the tree by choosing the nodes with the highest bound (\ref{eq:ucb}) until a leaf is reached.
As a further mechanism to promote exploration we add a small amount of Dirichlet noise to the policy.
When a terminal node is found, it is directly evaluated instead of being processed by the neural network.  This happens in two cases: when a maximum depth is reached, and when the special \texttt{STOP} operation is selected.
In the first case the search fails and the return is just zero.
In the second case the return depends on the distance between the resulting image $\vec{x}$ and the target image $\vec{y}$:
\begin{equation}
    \label{eq:return}
    r(\vec{x}) = \exp(-\alpha \| \vec{x} - \vec{y} \|^2),
\end{equation}
where $\alpha$ is a tunable parameter.

\begin{figure}[t]
\centering
\includegraphics[width=\columnwidth]{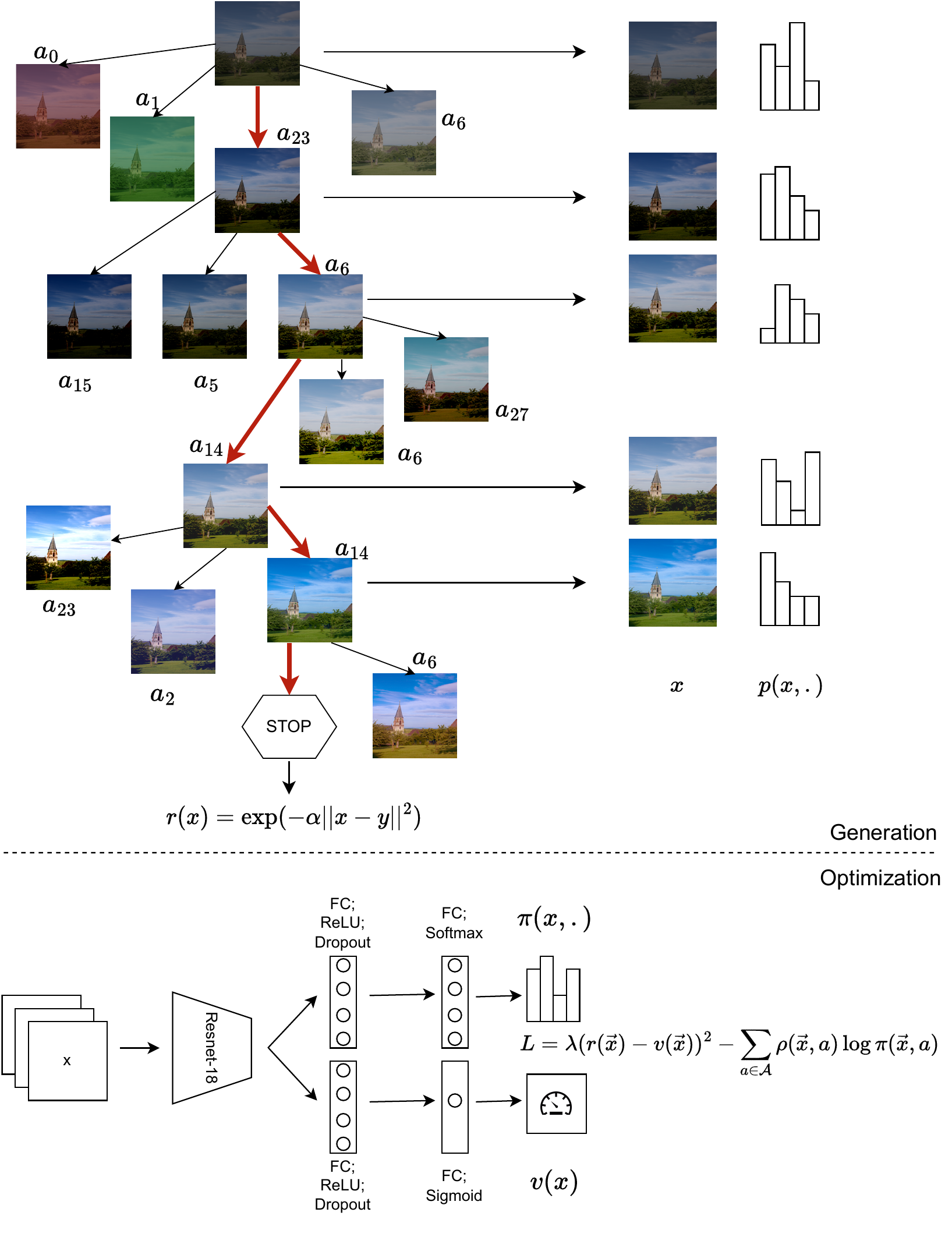}
\caption{\method{} generation (top) and optimization (bottom) phases. To make the figure more clear only part of the tree is actually shown. For the complete architecture of the ResNet-18, please refer to the original paper. The function $\pi(\vec{x}, \cdot)$ is the policy over the editing operations and $v(\vec{x})$ is the estimated return.}
\label{fig:pipeline}
\end{figure}

When a set number of MCTS iteration is reached, a policy $\rho$ for the root node is obtained.
\begin{equation}
    \rho(\vec{x}, a) = \frac{N(a(\vec{x}))}{N(\vec{x})}.
\end{equation}
This policy represents the probability distribution of the root's childs visits. From this probability distribution a random sample is extracted and the corresponding action is applied to the root image. This ensures a good degree of exploration by allowing the generation of multiple sequences of operations for the same image. From this new image (that becomes the new root of the tree) the complete procedure restarts and it is repeated until a terminal node is reached. The upper part of Figure \ref{fig:pipeline} depicts the generation phase of \method{}.

\subsubsection{Optimization}
The roots visited during the generation phase are used to build a training set.
Each training element is a triplet combining (i) the image $\vec{x}$, (ii) the return $r(\vec{x})$, and (iii) the probabilities of the stochastic policy $\rho(\vec{x}, \cdot)$. The parameters of the neural architecture are updated by minimizing the loss function $L$:
\begin{equation}
    \label{eq:loss}
    L = \lambda (r(\vec{x}) - v(\vec{x}))^2 -
    \sum_{a \in \mathcal{A}} \rho(\vec{x}, a) \log \pi(\vec{x}, a),
\end{equation}
which combines the squared estimation error of the return with the cross entropy measuring the disagreement between the stochastic policy $\rho$ and the network policy $\pi$. The parameter $\lambda$ is a tunable hyperparameter used to balance the weight of the two terms in the loss.

The whole process is repeated multiple times, each time producing more training data in the generation phase and improving the neural network in the optimization phase. 

\subsection{Neural Architecture}

The convolutional neural network used in this work 
is a modified version of the ResNet-18 where the last fully connected classification layer is replaced by two feed forward heads. Both include two fully connected linear layers. The first head also includes the softmax function to produce the vector representing the policy $\pi(\vec{x}, \cdot)$. The second one is followed by the application of the sigmoid activation function to compute the scalar value $v(\vec{x})$.   

We decided to consider the ResNet-18 as a backbone architecture 
since with a deeper model there is a high chance to overfit the (relatively small) datasets we used in the experiments. However, in principle there are no special limitations on the architecture of the neural network. 
The network is trained from scratch after random initialization.

The neural network accepts images of any spatial resolution.  The only limitation is  the amount of memory available in the system.
During training an image needs to be stored for each node in the tree.  Images can be downsampled to increase the number of trees that are grown in parallel and to speed-up training in general (in the experiments we resampled images to the resolution of $256 \times 256$ pixels).  
%

The Lower part of Figure \ref{fig:pipeline} shows the neural network architecture.  In the first layer of the two heads, in order to avoid overfitting, dropout is applied with probability $p$, where $p$ is a tunable hyperparameter. For our experiments the value of this parameter was tuned to 0.6.
%
%


\subsection{Inference}
\label{inference}
After training, \method{} can be used in two different ways: by building a tree (tree search) or by using directly the trained neural network (policy-based search).
The first strategy is more accurate but it also slower. The second is faster and slightly less precise.

\subsubsection{Tree Search}
at inference time the method builds the tree as in the generation phase, but without having access to the target image for the computation of the return of terminal nodes.
When the \texttt{STOP} action is selected or the maximum depth is reached the return is estimated instead by $v(\vec{x})$.
Once the tree has been built, a sequence of operations is obtained by descending the tree taking at each node the actions that maximize the policy $\rho(\vec x, \cdot)$.

\subsubsection{Policy-based Search}
the second approach consists in the repeated selection of the operation that by maximizes the probability $\pi(\vec x,\cdot)$ computed by the neural network.  The procedure terminates when the \texttt{STOP} action is selected or the maximum number of operations is reached.

\section{Experimental Results}
\label{sec:exp_res}
To verify the quality of the solutions found by \method{} we evaluated it on two datasets widely used to assess the performance of image enhancement methods.  More in detail, we used the LOw-light (LOL) dataset and the Adobe five-K dataset.  For both datasets global editing operations for tone adjustment and contrast enhancement are required.  In the case of LOL spatial operators (such as deblurring, edge enhancement, etc.) can also be useful.

\subsection{LOw-Light Dataset}
\label{subsec:LOL}
The LOL dataset \cite{Chen2018Retinex} is composed of 500 pairs of low-light/normal-light images obtained from real scenes. The training set includes 485 images and the test set 15. One of the main problems of real-scene datasets is the alignment of the image pairs: for this reason the authors of the dataset removed all the pairs that have a mean squared error greater than a fixed threshold (0.1).
The enhancement of the images in this dataset requires mostly the application of global transformations,
white balancing algorithms and spatial filters.  The set of operations $\mathcal{A}$ we selected for this experiment includes:
\begin{itemize}
    \item \emph{Brightness adjustment}: a constant $\delta$ is added to the values of all pixels:
    \begin{equation}
        (a(\vec{x}))_{ijc} = \operatorname{clip}(\vec{x}_{ijc} + \delta).
    \end{equation}
    The $\operatorname{clip}$ function limits the adjusted values to the $[0, 1]$ range ($\operatorname{clip}(x) = \min \{\max\{ x, 0 \}, 1\}$).
    We considered eight variants of this operator:
    with $\delta \in \{-0.1, +0.1\}$, and applied to all the color channels or to a single color channel at a time $c \in \{R, G, B\}$.  
    \item \emph{Contrast adjustment}: the pixels values are stretched by a factor $\beta$ around their mean:
    \begin{equation}
        (a(\vec{x}))_{ijc} = \operatorname{clip}(\mu_c + \beta \times (\vec{x}_{ijc} - \mu_c)),
    \end{equation}
    where $\mu_c$ is the average channel value. We considered the variants with $\beta \in \{0.8, 2.0\}$ and applied to all the channels, or just one.
    \item \emph{Gamma Correction}: we considered two values $\gamma \in \{0.6, 1.1\}$, and applied them to all or to one of the channels:
    \begin{equation}
        (a(\vec{x}))_{ijc} = (\vec{x}_{ijc})^\gamma.
    \end{equation}
    \item \emph{Saturation adjustment}: after the conversion to the HSV color space, the $S$ channel is scaled by a factor $s \in \{0.5, 2.0\}$ and then the pixel is converted back in the RGB space.
    \item \emph{Hue rotation}; the hue channel of the HSV color space is shifted by $h$ degrees ($h \in \{-18.0, +18.0\}$).
    
    \item \emph{Gray world}: the gray world white balancing algorithm is applied to the image.  This algorithm makes the assumption that the images, on average, are gray and the value of the illuminant (in each channel) could be modeled as a constant factor multiplied to the channel. For this reason, in order to remove the variation due to the illuminant, the algorithm normalizes each channel of the image with the average value of the pixels. 
    \item \emph{Max RGB}: the max RGB white balancing algorithm is applied.  This algorithm relies on the fact that the brightest pixel in each channel should be considered as white. Each channel is divided by the highest value. 
    \item \emph{Median Filter}: replaces the value of a pixel with the median of the values of its neighbours. The dimension $k$ of the filter's window was tune to 3 obtaining a $3 \times 3$ filter.  
    \item \emph{Gaussian Blur Filter}: blurs the input image with a Gaussian filter with standard deviation $\sigma$. The parameter $\sigma$ was set to 2.  
        \item \emph{Sharpening Filter}: increases the sharpness of intensity transitions in an image. The sharpening filter that we considered is the unsharp masking, defined by the kernel $F$:
        \begin{equation} F=
        \begin{bmatrix}
            -0.125 & -0.125 & -0.125\\
            -0.125 & 2 & -0.125\\
            -0.125 & -0.125 & -0.125
        \end{bmatrix}.
        \end{equation}
        \item \emph{Edge Enhancement Filter}: highlights the edges in the image.
        \begin{equation} F=
        \begin{bmatrix}
            -0.5 & -0.5 & -0.5\\
            -0.5 & 5 & -0.5\\
            -0.5 & -0.5 & -0.5
        \end{bmatrix}.
        \end{equation}
        \item \emph{Detail Filter}: the main purpose of this filter is to enhance the details of the subjects/objects present in the image. 
        \begin{equation}F =
        \begin{bmatrix}
            0 & -0.17 & 0\\
            -0.17 & 1.67 & -0.17\\
            0 & -0.17 & 0
        \end{bmatrix}.
        \end{equation}
        \item \emph{Smoothing Filter}: replaces the value of the pixels with the average of their neighbors in the filter window.  
        \begin{equation}F =
        \begin{bmatrix}
            0.077 & 0.077 & 0.077\\
            0.077 & 0.385 & 0.077\\
            0.077 & 0.077 & 0.077
        \end{bmatrix}.
        \end{equation}
        
    \item \emph{\texttt{STOP}}: terminates the enhancement procedure.
\end{itemize}
The total number of editing operations is 38 (eight brightness adjustments, eight contrast adjustments, eight gamma corrections, two saturation adjustments, two hue rotations, two white balancing algorithms, seven spatial filters and one stop). The parameters $\delta, \beta, \gamma, s, h, \sigma, k$ have been chosen to make it possible to enhance the images within a reasonable amount of steps. In the case in which, to save memory, \method{} is applied to resampled images, the parameters of the spatial operators need to be adjusted when finally applied to the original high resolution images.

The parameters of the training procedure described in Section~\ref{sec:method} have been selected on the basis of preliminary experiments, also taking into account the cost in terms of computational resources.  In each generation phase we randomly selected \num{100} training images and for each one we applied \num{10000} MCTS iterations to grow the corresponding tree (we set to ten the maximum tree depth).
The resolution of images was $600 \times 400$ pixels and standard data augmentation techniques (cropping, flipping and resizing) were used to improve the diversity of the training set.
Ten $(\vec{x}, r(\vec{x}), \rho(\vec{x}, \cdot))$ triplets were sampled from each tree and added to the training set. The parameter $\alpha$ in Equation (\ref{eq:return}) was set to 0.05 and the exploration parameter $c$ in Equation (\ref{eq:ucb}) was set to $10$.  

In each optimization phase the parameters of the neural network have been updated by minimizing the average loss (\ref{eq:loss}) with 160 iterations of the ADAMW
 optimizer, with learning rate $10^{-3}$ and weight decay $10^{-2}$.

At the end of the training procedure the method has been applied to the test set . For each test image a tree of \num{1000} nodes has been grown and a final editing sequence was selected with the procedures described in Section~\ref{inference}. Table \ref{tab:res_lol} reports the result of \method{} on the LOL test set with the Tree Search inference procedure.

We compared the results of our method, with several state of the art architectures in terms of PSNR, SSIM, $\Delta E$.   
\begin{table*}[t]
    \caption{Comparison of state-of-the-art enhancement methods on the LoL dataset.}
    \makebox[\textwidth]{\resizebox{\linewidth}{!}{\def\arraystretch{1.5}%
    \begin{tabular}{lcccccccccccccc}
    \toprule
         & \rotatebox{90}{RRDNet~\cite{zhu2020zero}}& \rotatebox{90}{DSLR~\cite{lim2020dslr}} & \rotatebox{90}{ExCNet~\cite{zhang2019zero}} & \rotatebox{90}{Zero-DCE~\cite{guo2020zero}} & \rotatebox{90}{RetinexNet~\cite{Chen2018Retinex}} & \rotatebox{90}{MBLLEN~\cite{lv2018mbllen}} & \rotatebox{90}{KinD~\cite{zhang2019kindling}} & \rotatebox{90}{LAE-Net~\cite{LIU2022109039}}&\rotatebox{90}{KinD++~\cite{zhang2019kindling}} & \rotatebox{90}{EnlightenGAN~\cite{jiang2021enlightengan}} & \rotatebox{90}{DPED~\cite{ignatov2017dslr}} & \rotatebox{90}{HDRNet~\cite{gharbi2017deep}} &\rotatebox{90}{RetinexDec~\cite{lv2021low}}& \rotatebox{90}{\textbf{TreEnhance}} \\
    \midrule
         PSNR$\uparrow$&11.36&15.23&15.80&16.15&17.16&17.38& 18.27& 18.30& 18.75& 18.75& 19.71& 20.14& 20.23&\textbf{21.96}\\
         SSIM$\uparrow$&0.54&0.68&0.67&0.70&0.72&0.75&0.83& 0.64&0.82&0.78& 0.81& 0.83&\textbf{0.84}& 0.81\\
         $\Delta E\downarrow$&32.63&23.47&21.47&21.99&18.31&20.48&15.02&-&15.13&15.06& 14.81& 14.26&13.10 & \textbf{10.82} \\
    \bottomrule
    \end{tabular}}}
    \label{tab:res_lol}
\end{table*}

\begin{figure}[t]
    \centering
        \includegraphics[scale=0.465]{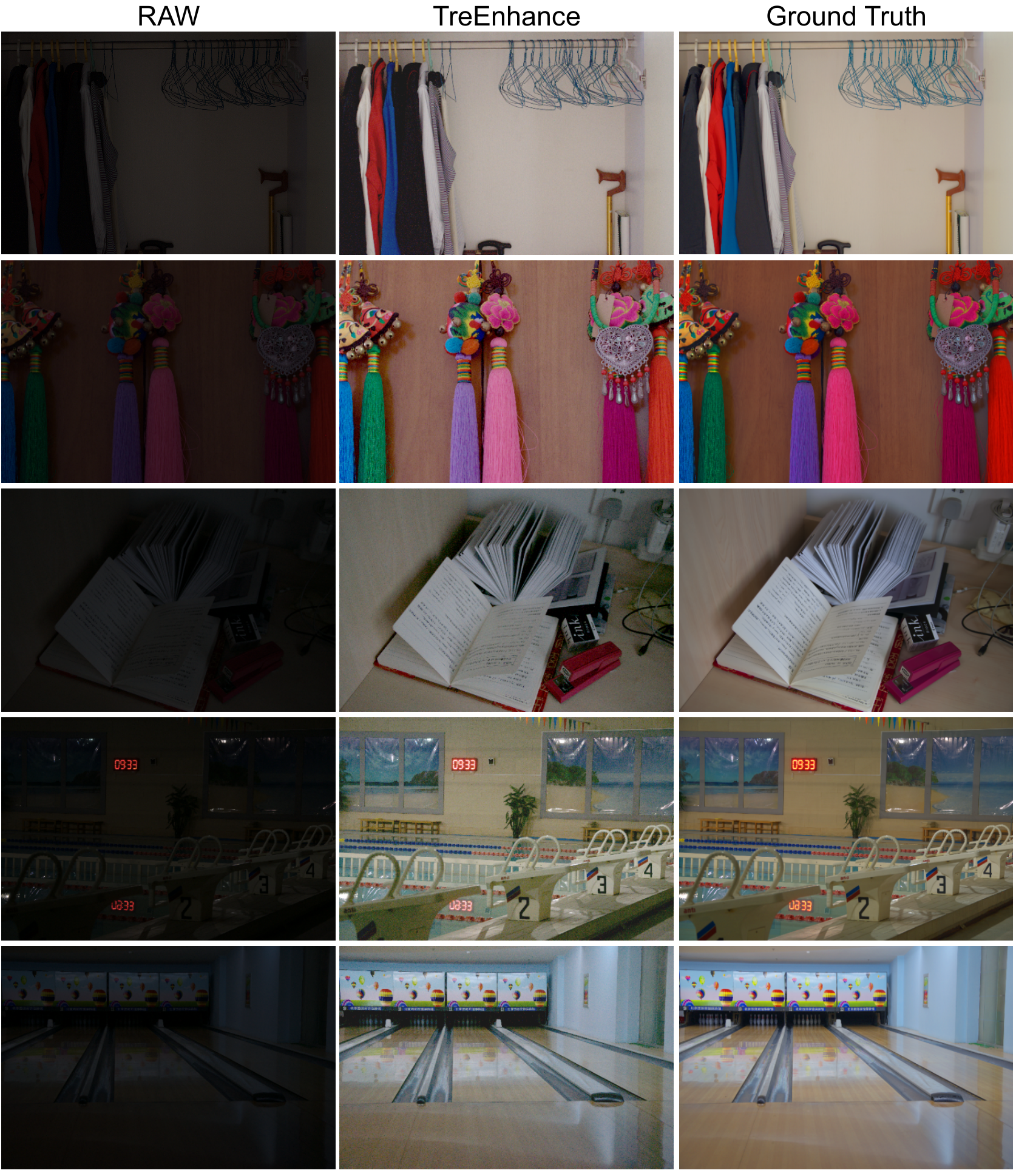}
    \caption{Results of the application of \method{} on the LOL test set. From left to right: low-light images, \method{} output and ground truth images.}
    \label{fig:lolresults}
\end{figure}

\method{} was able to outperform all the other methods considered in terms of PSNR and $\Delta E$.
Figure \ref{fig:lolresults} presents the results of the application of \method{} on the LOL test set. From a qualitative and a quantitative point of view the results obtained are very satisfactory. 

\subsection{Adobe Five-K Dataset}
The second dataset we considered is the Adobe Five-K dataset \cite{fivek}. It includes \num{5000} high-resolution images in the device RAW format and in the Adobe RGB space after the work of five different photo editing experts (Expert A, B, C, D, E). We followed the procedure presented by Hu et al.~\cite{hu2018exposure} to split the images in a training (\num{4000} images) and a test set (\num{1000} images).  As target images we took those edited by Expert C, since it is the most consistent of the five.

For this second dataset, the set of enhancing operators $\mathcal{A}$ was reduced by removing the spatial filters and the white balancing algorithms, exploiting the knowledge that the experts used only those operations that can be obtained with the ``color curves'' tool of photo editing applications.
For the Five-K dataset, the difference between input and ground truth is smaller than that of the LOL dataset.
For this reason we changed the parameters of the operators obtaining more fine-grained actions.
\begin{itemize}
    \item \emph{Brightness adjustment}: $\delta \in \{-0.05, +0.05\}$;
    \item \emph{Contrast adjustment}: $\beta \in \{0.894, 1.414\}$; 
    \item \emph{Gamma Correction}: $\gamma \in \{0.775, 1.05\}$; 
    \item \emph{Saturation adjustment}: $s \in \{0.707, 1.414\}$; 
    \item \emph{Hue rotation}; $h \in \{-9.0, +9.0\}$.
\end{itemize}
The total number of operations is 29: the \texttt{STOP} action, two saturation adjustments, two hue rotations, brightness, contrast and gamma correction applied either to a single channel or to three channels simultaneously.

The training procedure adopted for this second dataset was the same explained in Section \ref{sec:method}. For each generation we considered 100 images from the training set, we downsampled them to $256 \times 256$ pixels and we applied data augmentation as for the LOL dataset. We executed \num{10000} MCTS iterations (the maximum depth was set to ten).
The value of the parameters $c$ in Equation (\ref{eq:ucb}) was set to 4: a value smaller than that used for the LOL dataset to account for the reduced number of operations that requires a smaller degree of exploration. The value of the parameter $\alpha$ in Equation (\ref{eq:ucb}) is 0.05, the same used for the LOL dataset.
After each generation phase, we trained the network with mini batch gradient descent using the ADAMW optimizer for 70 iterations. 

We compared our method (with the Tree Search inference procedure) with several other methods from the state of the art: results are reported in Table \ref{tab:results}.

\begin{table}[t]
    \caption{Comparison of state-of-the-art enhancement methods on the Adobe five-k dataset.}
    \makebox[\columnwidth]{\resizebox{\linewidth}{!}{\def\arraystretch{1.5}%
    \begin{tabular}{lccccccccccc}
    \toprule
        &\rotatebox{90}{Exposure~\cite{hu2018exposure}}&\rotatebox{90}{CycleGan~\cite{zhu2017unpaired}}&\rotatebox{90}{DCE-Net~\cite{zhang2021star}}&\rotatebox{90}{DCE-Net-Pooling~\cite{zhang2021star}}&\rotatebox{90}{DaR~\cite{park2018distort}}&\rotatebox{90}{Unfiltering~\cite{bianco2017artistic}}&\rotatebox{90}{Pix2Pix~\cite{isola2017image}}&\rotatebox{90}{HDRNet~\cite{gharbi2017deep}}&\rotatebox{90}{Star-DCE~\cite{zhang2021star}}&\rotatebox{90}{Parametric~\cite{bianco2019learning}}&\rotatebox{90}{\textbf{TreEnhance}}\\
    \midrule
        LPIPS$\downarrow$&\num{0.16}& \num{0.16}& \num{0.13}&$\num{0.15}$&\num{0.10} & \num{0.09}& \num{0.09}& \num{0.08}& $ \num{0.08}$& \num{0.07}& \textbf{0.06}\\
        PSNR$\uparrow$& \num{18.74}& \num{19.38}& 20.47& 20.33& \num{20.91}&\num{21.67}& \num{23.05}& \num{22.31}& \textbf{23.55}&\num{23.20}&\num{21.24}\\
        $\Delta E\downarrow$& \num{13.57}& \num{12.95}& $\num{11.77}$& $\num{12.13}$& \num{13.05}& \num{21.67}& \num{8.84}& \num{9.53}& $\textbf{8.45}$& \num{8.52} & \num{11.25}\\
        SSIM$\uparrow$& \num{0.81}& \num{0.78}& \num{0.85}& \num{0.85}& \num{0.86} & \num{0.88}& \num{0.86}& \num{0.89}& \num{0.89}& \textbf{0.90}& \num{0.89}\\
    \bottomrule
    \end{tabular}}}
    \label{tab:results}
\end{table}

\begin{figure}
    \centering
    \includegraphics[width=\columnwidth]{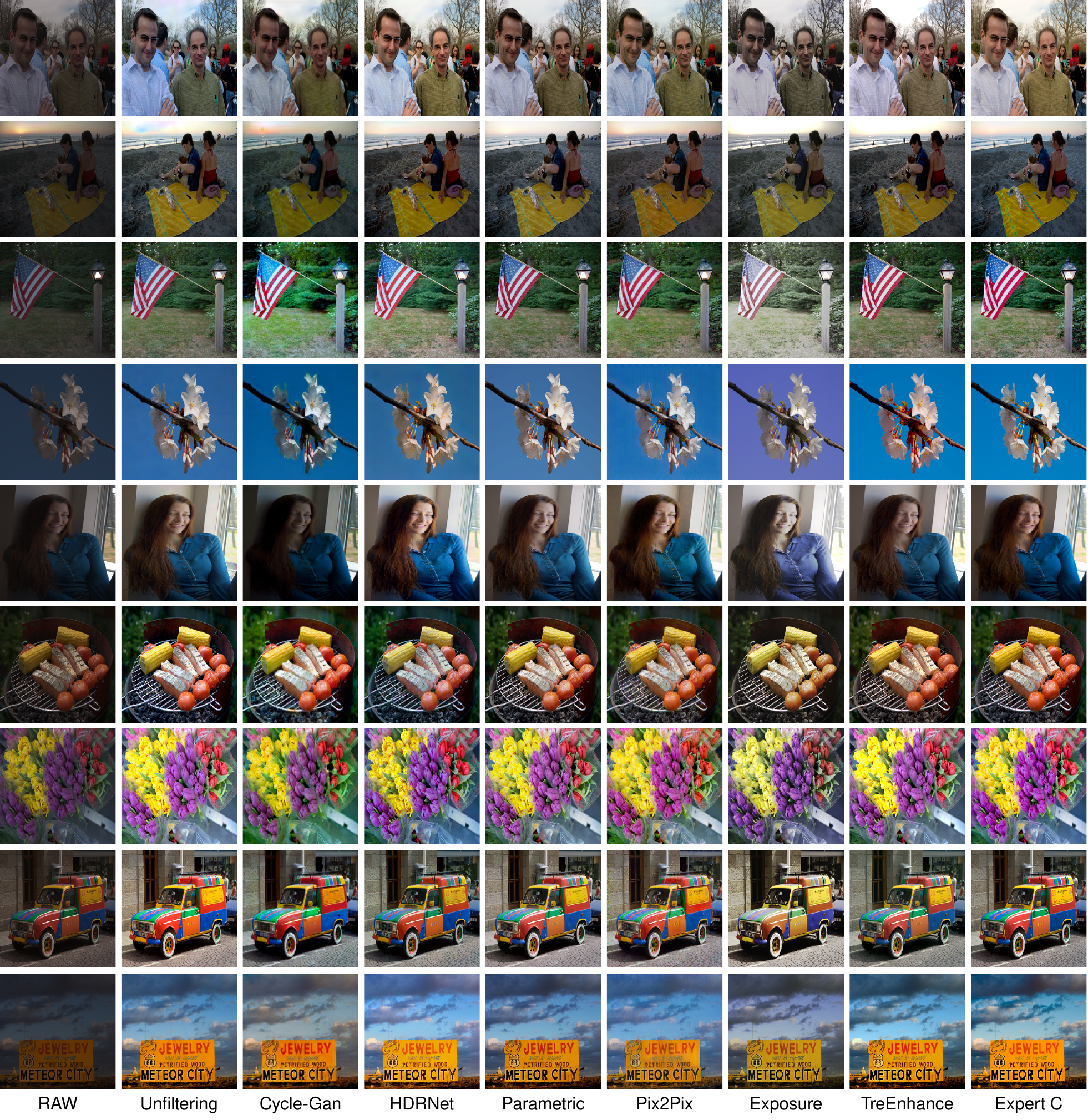}
    \caption{Examples of test images in the Five-K dataset processed by methods in the state of the art. The left most column shows the input image and the right most the ground truth.}
    \label{fig:comparison}
\end{figure}

Since the images provided in the Five-K are RAW format images, before training the algorithm, a preprocessing was needed. In order to obtain a fair comparison with the other state of the art methods, we applied these algorithms, using the code provided by the authors, to our preprocessed data.
From the comparison we can see that \method{} is competitive on this dataset performing very well from a quantitative point of view. Moreover, it reached the best results in terms of LPIPS~\cite{zhang2018perceptual} a metric that measures the perceptual similarity between two images. It also obtained a very good SSIM value.  In terms of the pixel-level metrics PSNR and $\Delta E$, other methods obtained better  results.  This fact depends on the granularity of the set of editing operations considered.  To obtain more accurate results at the pixel level, we would need to include more fine-grained operations, which would make significantly longer the sequences found by the algorithm. Moreover, when the sequences of operators becomes to long, the level of explainability of the model is reduced.
In Figure~\ref{fig:comparison} we compared the images obtained using \method{} and other state-of-the-art methods to enhance a sample from the Five-K test set.

\method{} is able to obtain very good performance on the two datasets considered. Its results are competitive for all the metrics considered, as reported in tables \ref{tab:res_lol} and \ref{tab:results}. However, it was not the best method on the Five-K dataset in terms of MSE and PSNR. A possible explanation for this is that
some Five-K images have been retouched with spatially varying operators (e.g. masked operations), a condition that is difficult to reproduce exactly by using only global operators.

To better understand the differences among the images produced by the methods in the comparison, we organized a subjective evaluation test.
We asked ten participants (all with some experience in the field) to select the best looking among seven versions of the same image. This selection process was repeated for 100 different images from the Five-K test set. The compared methods were \method{}, Pix2Pix, CycleGAN, Star-DCE, DaR, Exposure and HDRNet.  The participants used a system that shows the seven alternatives in random order that, without any reference to the methods.  The order used to present the 100 images was also different for each user.

\begin{figure}
    \centering
    \includegraphics[width=\columnwidth]{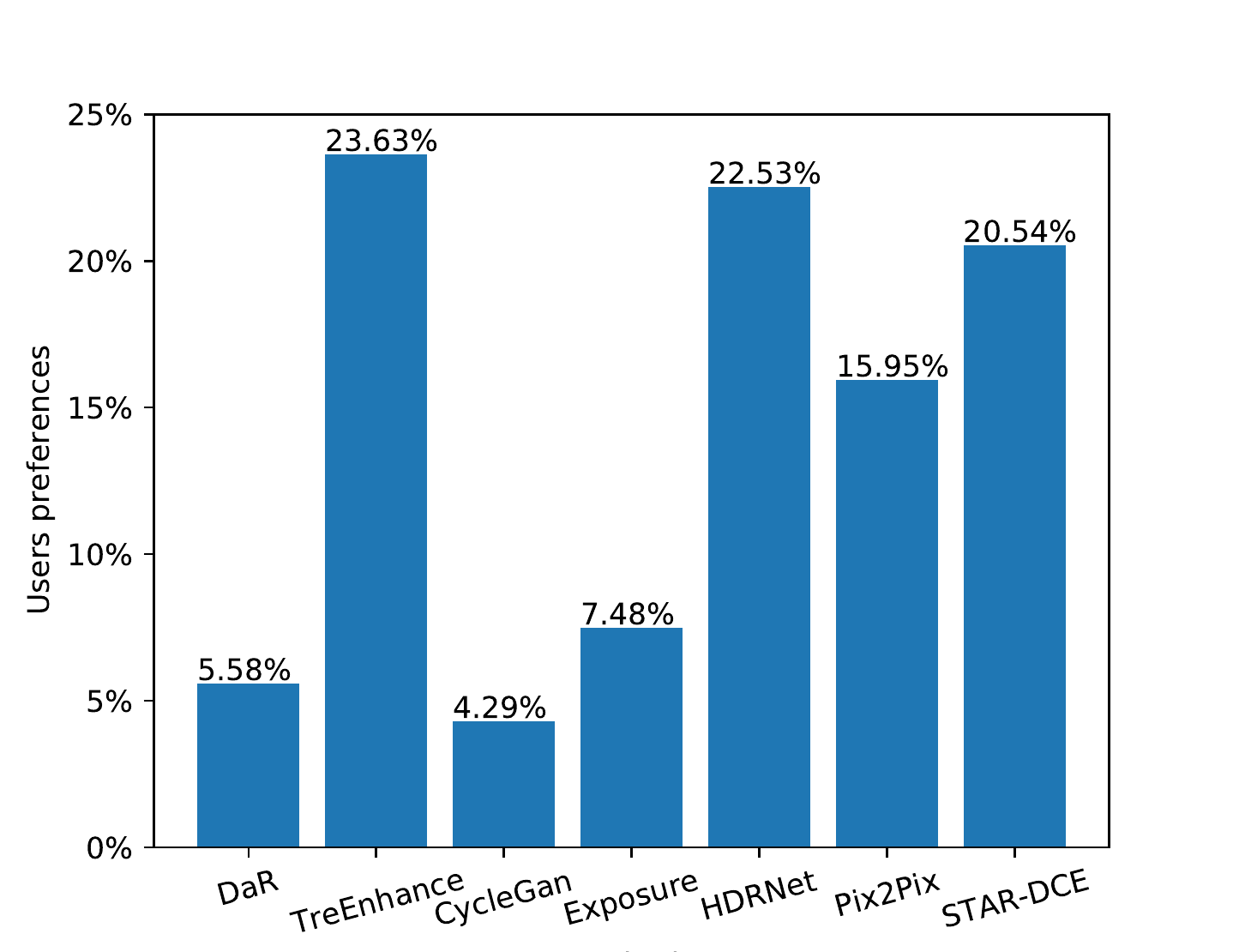}
    \caption{Results of the user study on the 100 images from the Five-K test set.}
    \label{fig:user_study}
\end{figure}

From the results shown in Figure \ref{fig:user_study}, \method{} was the most selected method with the \textbf{23.63\%} of the users preferences followed by HDRNet (22.53\%) and Star-DCE(20.54\%). The other methods showed much lower preferences: Pix2Pix 15.95\%, Exposure 7.48\%, DaR 5.58\% and CycleGAN 4.29\%.

These results confirm the ability of \method{} of producing high quality images from a perceptive point of view even if the pixel-level metrics are not the best one. This is confirmed by the best value of the perceptual LPIPS metric.
An informal interview after the test pointed out that for most users there were three methods that were clearly superior than the others. These methods, were very stable in the choice of the tonality and their output was free from any artifacts, a very negative element for the others.
The users considered the best three methods as substantially equivalent.

From the performed user study, \method{} is very competitive respect several methods for image enhancement.  Moreover is able to produce images similar to modern state-of-the-art methods.

In addition to this experiments, we also conducted exploratory  tests with probabilistic denoising diffusion models \cite{saharia2022palette}. Preliminary results show that these models are very good at preserving the semantic content of images without the introduction of artifacts. However, diffusion models were not able to emulate the intent of the photor editor, and the result is a low accuracy in terms of performance metrics (especially PSNR and $\Delta E$). A possible explanation of these results, is the limited dimension of the datasets used in our experiments compared to those used in the reference paper~\cite{saharia2022palette}.


%

\section{Discussion}
\label{sec:discuss}
The results obtained by \method{} in the two different tasks are satisfactory from both a quantitative and a qualitative point of view. The high quality output images are obtained with the application of basic image editing operations, making the process easy to explain.  During the inference phase \method{} evaluates several candidate sequences of operators but at the end only the most promising is selected and applied to the input image.

In this section we discuss the memory and time requirements of \method{}, and we analyze more in detail its qualities.

\subsection{Memory and Time Requirements}
\begin{figure*}
    \centering
        \includegraphics[scale=0.75]{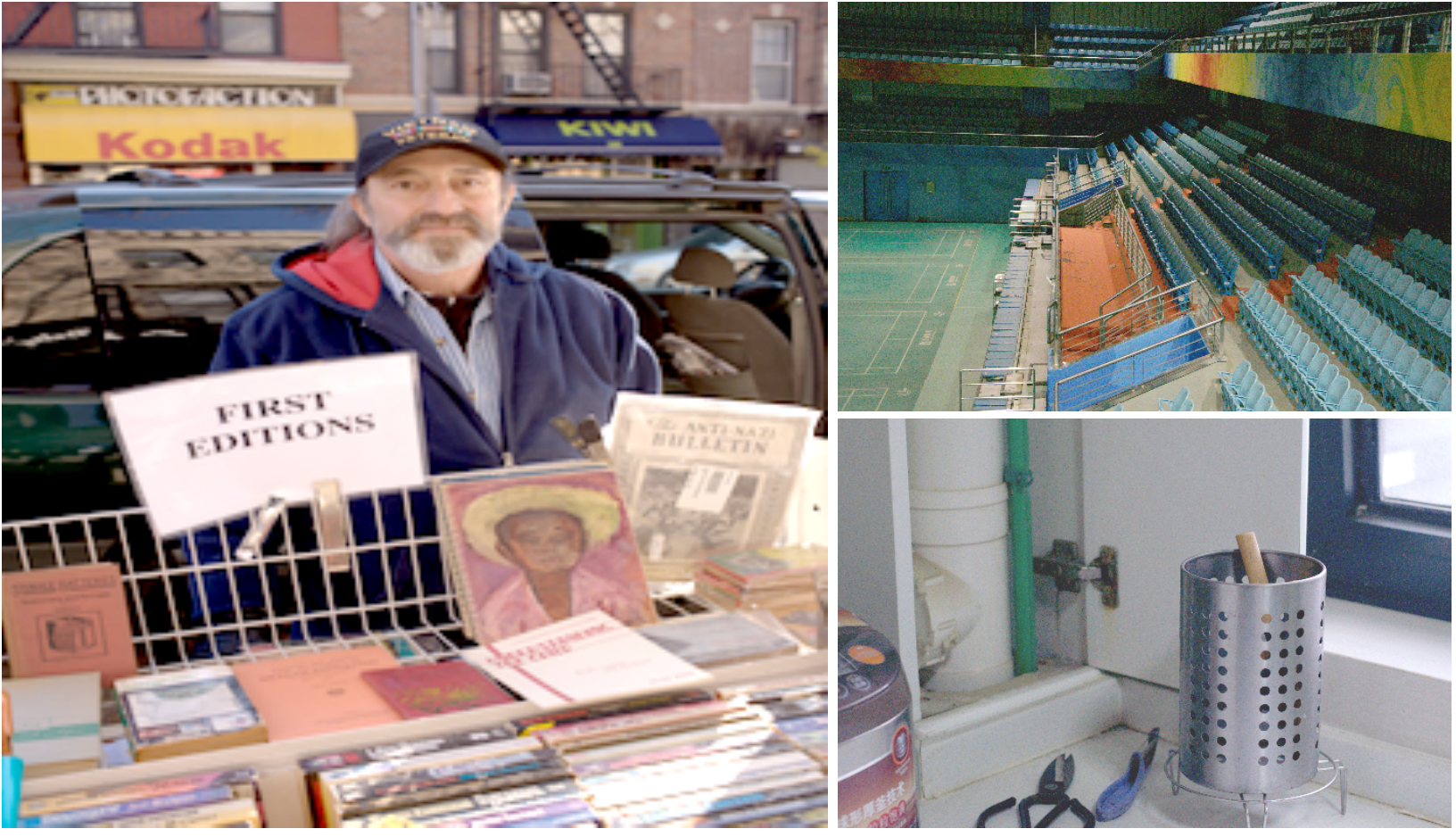}
    \caption{Examples of the application of \method{} to high resolution test images from the Five-k and LOL datasets.}
    \label{fig:result_256_fivek}
\end{figure*}

The generation phase of the training procedure is the most demanding step of \method{} in terms of memory. This depends on the number of MCTS  steps, which also corresponds to the number of nodes in the trees.
With respect to other state-of-the-art methods, \method{} does not impose any limitations on the resolution of the input images.  However, to save memory (and therefore to allow the growth of larger trees) it is possible to work with resampled images.
Once the sequence of editing operations has been found on a low-resolution version of the input image, it can be applied to the original high-resolution version.  In this case the parameters of spatial operators must be suitably adjusted.
%
Figure \ref{fig:result_256_fivek} shows some examples of the application of \method{} on high resolution images.

We observed that a good degree of exploration should be ensured in the initial phases of the training procedure. 
In fact, at the beginning it tends to look like a breadth-first search where many nodes at the same level are explored until a promising action is found.  After that, the method explores more in depth the promising branches of the tree.  This happens when the CNN becomes able to assign more accurate values allowing the algorithm to select promising actions more frequently. 

As explained in Section \ref{sec:method}, only a random subset of the training set is involved in each generation. This subset is split in batches (four in our experiments) in order to make use of all the available memory. To speed up the generation phase, the trees corresponding to each one of the images in the batch grow in parallel.  In our experiments we used an Intel\textregistered{} Core\texttrademark{} i7-8700 CPU @ 3.20GHz with 48 GB of memory.  For the evaluation step of the generation phase and for the optimization phase we used a NVIDIA RTX\texttrademark{} 3080Ti graphic card with 12 GB of memory. 

\subsection{Network Ablation}
In Section \ref{sec:method}, we presented the neural architecture used in this work. In Table~\ref{tab:ablation} we compared different neural network architectures in terms of PSNR on the LOL dataset.

\begin{table}[t]
    \caption{Comparison of different neural architectures on the LOL test set with the Tree-Based inference strategy.}
    \centering
    \resizebox{\columnwidth}{!}{\begin{tabular}{lccc}
    \toprule
         Architecture & Parameters (M) & FLOPs (G) & PSNR  \\
    \midrule
         ResNet18 & 11.32 & 2.38 & 21.96 \\
         ResNet34 & 21.79 & 4.80 & 20.43 \\
         ResNet50 & 25.56 & 5.38 & 19.30 \\
         ResNet101 & 44.55 & 10.52 & 17.13\\
         DenseNet121 & 7.98 & 3.76 & 20.13\\
    \bottomrule
    \end{tabular}}
    \label{tab:ablation}
\end{table}

We decided to compare two relatively small networks as Resnet18 and DenseNet121, two mid-size networks as Resnet34 and Resnet50 and the large ResNet101. As we stated before, the method does not impose any constraint for the choice of the neural architecture used to estimate the probability distribution over the actions and the outcome of the enhancement process. The ResNet18 and DenseNet121 have a similar number of parameters and FLOPs (computed for images of resolution 256 $\times$ 256). These networks show similar performance in terms of PSNR. Due to the relatively high number of forward passes required by the Tree-based strategy and to an higher PSNR obtained, we decided to select for our method the ResNet18 even if its number of parameters is higher than that of the DenseNet. The performance obtained with larger networks, are worse than those obtained with the smaller ResNet. This is caused by the overfitting of the relatively small training set.  We expect that even larger networks, such as ResNet152, would suffer the same problem.

\subsection{Tree-based and Policy-based Enhancement}
As explained in Section \ref{sec:method} \method{} can be used in the inference phase in two different ways: by building a tree for each input image or by directly using the neural network as a policy. In Table \ref{tab:method_onlypol} we compared the two different inference strategies on the two test sets. Tree Search enhancement is the most accurate strategy since it considers a wide range of alternatives.   Policy-based Search strategy acts \emph{greedy} by taking each time the most promising operation. For this reason it is faster than the tree-based approach and uses less memory but is slightly less accurate.
\begin{table}
    \caption{Comparison of \method{} inference strategies on the two dataset considered.}
    \centering
    \resizebox{\columnwidth}{!}{
    \begin{tabular}{lcccccc}
    \toprule
        & \multicolumn{3}{c}{Five-K} & \multicolumn{3}{c}{LOL} \\
        \cmidrule(l){2-4}
        \cmidrule(l){5-7}
        Search strategy & 
         PSNR & $\Delta E$ & SSIM & PSNR & $\Delta E$ & SSIM \\
         \midrule
         \emph{Tree} & 21.24 & 11.25 & 0.89 & 21.96 & 10.82 & 0.81 \\
         \emph{Policy-based} & 20.06 & 12.44&0.84 & 21.86&10.99 & 0.80 \\
         \emph{Guided} & 25.54 & 7.18 & 0.92 & 24.34 & 8.44 & 0.84 \\
         \bottomrule
    \end{tabular}
    }
    \label{tab:method_onlypol}
\end{table}

\begin{figure}
    \centering
    \begin{tabular}{cc}
    \includegraphics[scale=0.75]{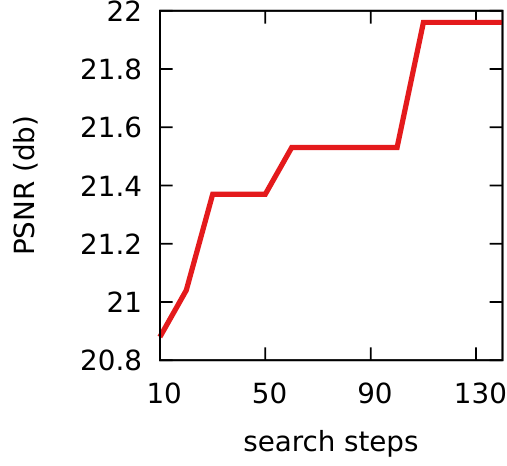} &
    \includegraphics[scale=0.75,trim=20 -3 0 0,clip]{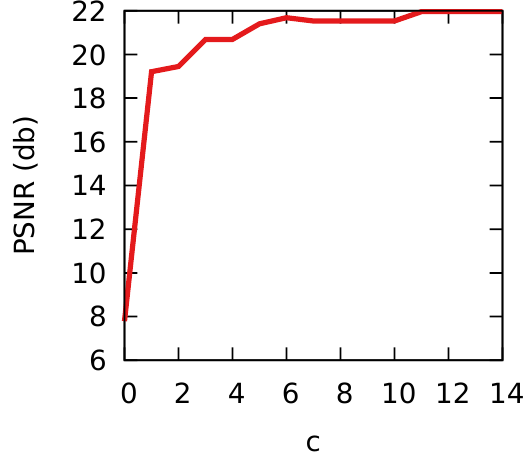} \\
    \end{tabular}
    \caption{Accuracy obtained by the \method{} on the LOL test set varying the number of MCTS search steps (left) and the exploration/exploitation coefficient (right).}
    \label{fig:inference-plots}
\end{figure}

The additional complexity of the Tree Search strategy can be modulated by limiting the number of MCTS steps during in the inference phase.  Figure~\ref{fig:inference-plots} (left) shows that the accuracy of the method increases with the number of steps.  Another parameter that can be adjusted during the search is the coefficient $c$ in Equation (\ref{eq:ucb}).  From Figure~\ref{fig:inference-plots} (right) it seems that during inference a large value of $c$ is beneficial.

We compared the two inference strategies in terms of time, number of parameters and FLOPs. Both the solutions are based on the modified version of the ResNet18 architecture presented before. The number of parameter is 11.32M. As explained in Section \ref{sec:method}, the main difference in the two inference phases is the fact that the first is growing the tree also during inference while the second directly applies the learned policy. The cost in terms of FLOPs for a forward pass on a 256 $\times$ 256 image is 2.38GFLOPs. For the tree strategy, the total computational cost depends on the number of times the network is evaluates an image associated to a node which, in turn, depends on the maximum number of steps and on the maximum depth of the tree. For the policy-based strategy, the image is evaluated by the neural network until the maximum number of editing operators have been applied or the \texttt{STOP} action has been selected.  In Table \ref{tab:cost} we compared the computational cost and time of the two search strategies on the Five-K test set with those of other methods from the literature. We computed the average FLOPs per image and the average time per image. Time was measured on a computer with a Intel\textregistered{} Core\texttrademark{} i7-8700 CPU @ 3.20GHz, 48 GB of memory and anNVIDIA RTX\texttrademark{} 3080Ti GPU with 12 GB of memory.
\begin{table}[t]
    \centering
    \resizebox{\columnwidth}{!}{\begin{tabular}{lcccc}
    \toprule
         Method & Params(M) &FLOPs(G)&Time(ms)&PSNR  \\
         \midrule
         \method{} (Tree) &11.32&40.54&397.14&21.24\\ 
         \method{} (Policy) &11.32&12.43&66.91&20.06\\
         CycleGAN &11.37 & 56.88 & 1642.95 & 19.38\\
         Pix2Pix & 54.41 & 18.15 & 1693.44 & 23.05 \\
         STAR-DCE & 0.03 & 0.02 & 25.17 & 23.55\\
         DCE-Net & 0.08 & 5.22 & 44.72 & 20.47\\
         HDRNet & 0.48 & 0.17 & 267.54 & 22.31\\
         Unfiltering & 148.02 & 2.59 & 816.56 & 21.67 \\
         Parametric & 0.03 & 0.05 & 756.93 & 23.20\\
         
    \bottomrule
    \end{tabular}}
    \caption{Comparison of the two inference strategies in terms of parameters, FLOPs and Time on the Five-K test set. Additionally, computational requirements of TreEnhance are also compared to those of other methods from the literature.
     }
    \label{tab:cost}
\end{table}

As expected, the Tree strategy is more accurate than the policy-based strategy, at the cost of a higher number FLOPs and, therefore, more time per image.

\subsection{Guided Search}

As a further experiment we verified if \method{} can be used as a tool for ``reverse engineering'' the work of a professional photo editor. To do so, we modified \method{} by using the ground truth during the inference phase (Equation \ref{eq:return}). This way the MCTS search will retrieve the sequence of operations that can be used to obtain a retouched version of the image from the input. We called this third strategy ``Guided Search''. The guided search is useful for two different reasons; moreover, it provides an upper bound to \method{} that can help the selection of the pool of enhancement operators for a new dataset; moreover, it could be included in a photo editing application, to help beginners to imitate expert users and to help professionals to revise and optimize their enhancement pipelines.

We reported the results obtained in Table \ref{tab:method_onlypol}.
The accuracy obtained in this scenario is very high confirming the effectiveness of MCTS in finding good sequences of image editing operations. This experiment suggests that this strategy could be included in an educational software, like a puzzle to be solved, to teach beginners how to enhance a low-quality image and what kind operators they should use to obtain a given result.

%

\subsection{Analysis of the Sequences}
\begin{figure}
    \centering
    \includegraphics[width=\columnwidth]{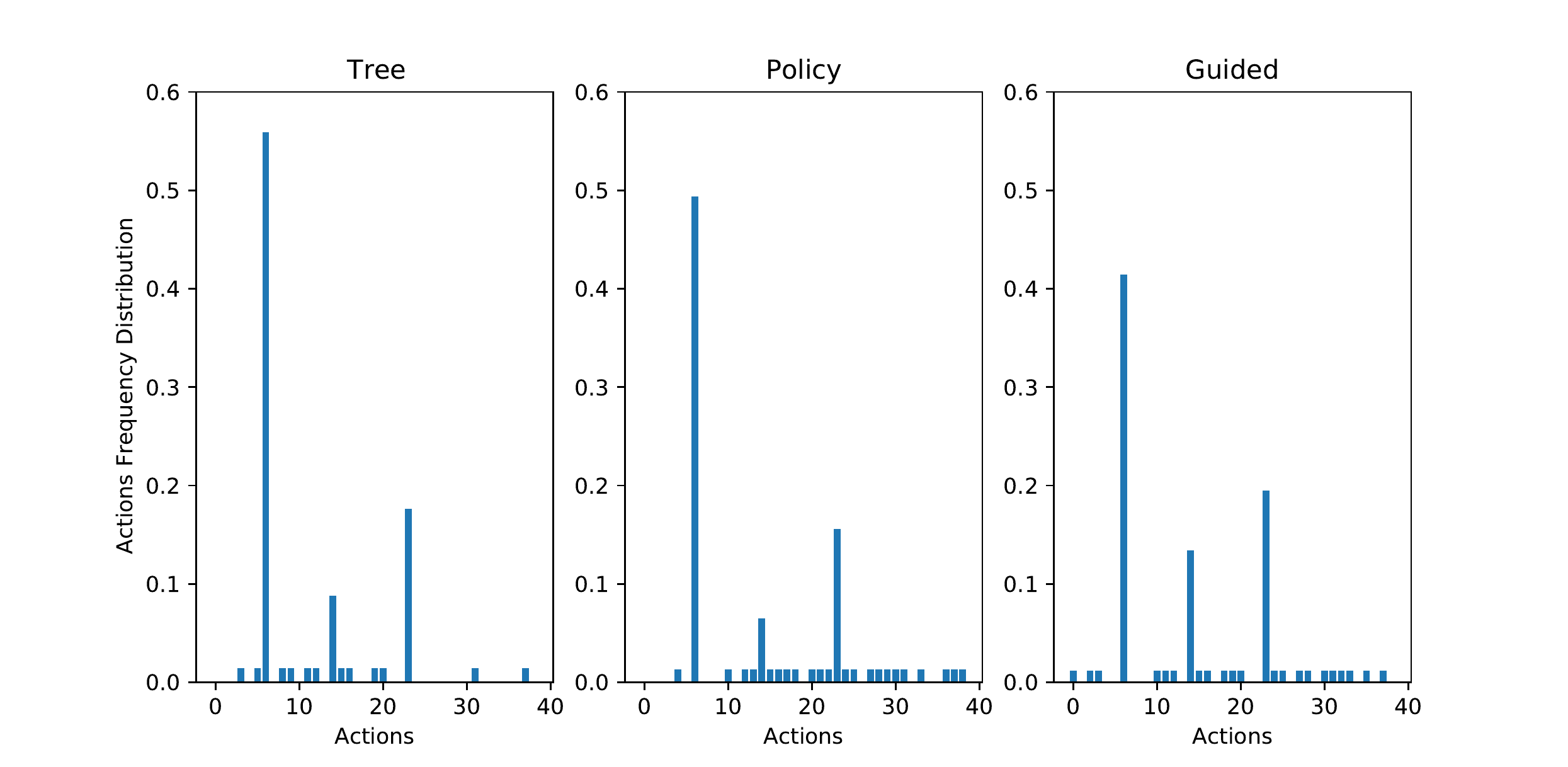}
    \caption{Frequency of the operations selected by the three inference strategies applied to the LOL test set. The action numbers reported here follows the order of presentation of subsection \ref{subsec:LOL}}
    \label{fig:seq_analysis}
\end{figure}

The sequences produced by the three inference strategies are  correlated. In fact, as it is shown in Figure \ref{fig:seq_analysis}, all the strategies favor global enhancing operators. In particular, the most selected operators are those that increase the brightness, the gamma, and the contrast. This depends on the fact that the original LOL test images are low light ones and need to be brightened.  Operations that work on single channels are more prone to be selected when the current image is close to the target one.

Tree-based and policy-based strategies tend to over-select the gamma correction action all over the tree channels with respect to the guided search (action 6). Guided search, in fact, tends to have a more balanced distribution of the operations over the three channels with respect to the other strategies, resulting in higher quality images.

\subsection{Artifacts}
\label{sec:artif}
\begin{figure*}[h]
    \centering
    \includegraphics[scale=0.33]{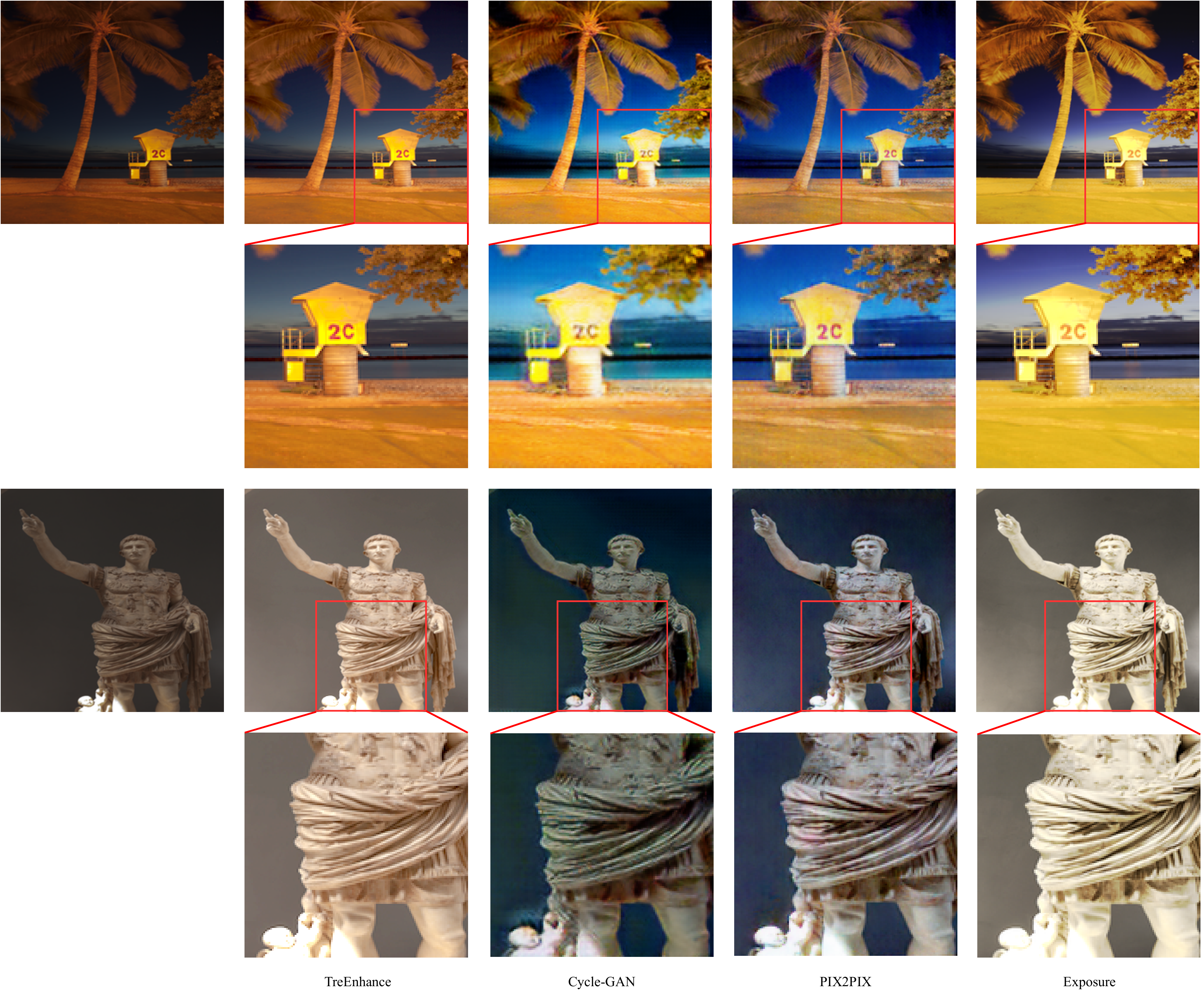}
    \caption{Comparison of generative and reinforcement learning methods on two Five-K test images.}
    \label{fig:artifacts}
\end{figure*}

One of the main features of the generative models for image enhancement, like those based on GANs, is the ability to accurately model color distributions.
However, as a drawback they often introduce visible artifacts in their outputs.
Figure~\ref{fig:artifacts} shows a comparison among two generative methods (Pix2Pix and CycleGAN) and two reinforcement learning methods (\method{} and Exposure) on two Five-K test images.  In both cases generative methods produced a very good color distribution, but with the presence of visible noise and artifacts.  In fact, the value of perceptual metrics such as LPIPS tends to be relatively bad for this kind of methods. Reinforcement learning methods do not introduce any artifacts, but at the cost, sometimes, of less saturated colors.

\section{Conclusions}
\label{sec:concl}

In this paper we presented \method{}, a light-weight fully automatic and explainable low-light image enhancement method. \method{} shows that it is possible to obtain very good performance in two image enhancement tasks.  It finds solutions that are fully explainable, since they can be decomposed as a sequence of simple image processing operations.

The method can be used for other enhancement tasks just by changing the set of image editing operations.  Thanks to the support for multiple search strategies, \method{} is suitable for a variety of scenarios with different requirements in terms of constraints on memory and processing time, and with different levels of interaction with the user.


More in detail experiments demonstrated that the method can obtain good performance on two low-light datasets with the possibility to choose three different search strategies: Tree-based, building the tree for each input image, and choosing the most promising path toward the enhanced image without using the ground truth image, Policy based, where the policy is sampled directly from the neural network, and Guided search, where the tree is created with the usage of the ground truth image.

The results obtained are very good from a qualitative and a quantitative points of view. However, \method{} shows some limitations when the enhancement style it is required to reproduce is not consistent, or when it requires complex non-global editing operations. To solve this issue, a more complex and dataset-specific set of editing operators should be designed, possibly by also including spatially varying operations.


\method{} demonstrates how tree-search theory, and in particular MCTS, can be a very useful tool for image processing. As a future expansion of this work, we plan to investigate the application of tree search theory in different research areas. One of these, could be the field of image retargeting where the MCTS could be able to select areas of the image that could be deleted, with the help of a series of retargeting operators, in order to center correctly the subjects of the image. The application of \method{} to other relevant fields, like super-resolution and image retrieval, could give the possibility of not only achieving great results with the application of basic operators, providing an explaination to the processing steps applied to the image.

Another promising direction of research is the design of diffusion models for low-light image enhancement.  In our preliminary tests, these generative models were able to produce very good images without artifacts, but they were not able to emulate the enhancement style of the photo editor. In the future, we plan to investigate how to overcome these limitations of diffusion models.

\bibliographystyle{elsarticle-num}
\bibliography{bibliography}

\begin{thebibliography}{10}
\expandafter\ifx\csname url\endcsname\relax
  \def\url#1{\texttt{#1}}\fi
\expandafter\ifx\csname urlprefix\endcsname\relax\def\urlprefix{URL }\fi
\expandafter\ifx\csname href\endcsname\relax
  \def\href#1#2{#2} \def\path#1{#1}\fi

\bibitem{isola2017image}
P.~Isola, J.-Y. Zhu, T.~Zhou, A.~A. Efros, Image-to-image translation with
  conditional adversarial networks, in: Proceedings of the IEEE conference on
  computer vision and pattern recognition, 2017, pp. 1125--1134.

\bibitem{zhu2017unpaired}
J.-Y. Zhu, T.~Park, P.~Isola, A.~A. Efros, Unpaired image-to-image translation
  using cycle-consistent adversarial networks, in: Proceedings of the IEEE
  international conference on computer vision, 2017, pp. 2223--2232.

\bibitem{ignatov2017dslr}
A.~Ignatov, N.~Kobyshev, R.~Timofte, K.~Vanhoey, L.~Van~Gool, Dslr-quality
  photos on mobile devices with deep convolutional networks, in: Proceedings of
  the IEEE International Conference on Computer Vision, 2017, pp. 3277--3285.

\bibitem{saharia2022palette}
C.~Saharia, W.~Chan, H.~Chang, C.~Lee, J.~Ho, T.~Salimans, D.~Fleet,
  M.~Norouzi, Palette: Image-to-image diffusion models, in: ACM SIGGRAPH 2022
  Conference Proceedings, 2022, pp. 1--10.

\bibitem{kim2022diffusionclip}
G.~Kim, T.~Kwon, J.~C. Ye, Diffusionclip: Text-guided diffusion models for
  robust image manipulation, in: Proceedings of the IEEE/CVF Conference on
  Computer Vision and Pattern Recognition, 2022, pp. 2426--2435.

\bibitem{sasaki2021unit}
H.~Sasaki, C.~G. Willcocks, T.~P. Breckon, Unit-ddpm: Unpaired image
  translation with denoising diffusion probabilistic models, arXiv preprint
  arXiv:2104.05358 (2021).

\bibitem{batzolis2021conditional}
G.~Batzolis, J.~Stanczuk, C.-B. Sch{\"o}nlieb, C.~Etmann, Conditional image
  generation with score-based diffusion models, arXiv preprint arXiv:2111.13606
  (2021).

\bibitem{lundberg2017unified}
S.~M. Lundberg, S.-I. Lee, A unified approach to interpreting model
  predictions, Advances in neural information processing systems 30 (2017).

\bibitem{selvaraju2017grad}
R.~R. Selvaraju, M.~Cogswell, A.~Das, R.~Vedantam, D.~Parikh, D.~Batra,
  Grad-cam: Visual explanations from deep networks via gradient-based
  localization, in: Proceedings of the IEEE international conference on
  computer vision, 2017, pp. 618--626.

\bibitem{ravirathinam2021c}
P.~Ravirathinam, D.~Goel, J.~J. Ranjani, C-lienet: A multi-context low-light
  image enhancement network, IEEE Access 9 (2021) 31053--31064.

\bibitem{jiang2021enlightengan}
Y.~Jiang, X.~Gong, D.~Liu, Y.~Cheng, C.~Fang, X.~Shen, J.~Yang, P.~Zhou,
  Z.~Wang, Enlightengan: Deep light enhancement without paired supervision,
  IEEE Transactions on Image Processing 30 (2021) 2340--2349.

\bibitem{lv2018mbllen}
F.~Lv, F.~Lu, J.~Wu, C.~Lim, {MBLLEN}: Low-light image/video enhancement using
  {CNNs}., in: BMVC, 2018, p. 220.

\bibitem{lore2017llnet}
K.~G. Lore, A.~Akintayo, S.~Sarkar, Llnet: A deep autoencoder approach to
  natural low-light image enhancement, Pattern Recognition 61 (2017) 650--662.

\bibitem{zhang2019kindling}
Y.~Zhang, J.~Zhang, X.~Guo, Kindling the darkness: A practical low-light image
  enhancer, in: Proceedings of the 27th {ACM} international conference on
  multimedia, 2019, pp. 1632--1640.

\bibitem{lv2021low}
X.~Lv, Y.~Sun, J.~Zhang, F.~Jiang, S.~Zhang, Low-light image enhancement via
  deep retinex decomposition and bilateral learning, Signal Processing: Image
  Communication 99 (2021) 116466.

\bibitem{zhu2020zero}
A.~Zhu, L.~Zhang, Y.~Shen, Y.~Ma, S.~Zhao, Y.~Zhou, Zero-shot restoration of
  underexposed images via robust retinex decomposition, in: {IEEE}
  International Conference on Multimedia and Expo ({ICME}), 2020, pp. 1--6.

\bibitem{cai2021learning}
Y.~Cai, X.~Hu, H.~Wang, Y.~Zhang, H.~Pfister, D.~Wei, Learning to generate
  realistic noisy images via pixel-level noise-aware adversarial training,
  Advances in Neural Information Processing Systems 34 (2021) 3259--3270.

\bibitem{dhariwal2021diffusion}
P.~Dhariwal, A.~Nichol, Diffusion models beat gans on image synthesis, Advances
  in Neural Information Processing Systems 34 (2021) 8780--8794.

\bibitem{bianco2017artistic}
S.~Bianco, C.~Cusano, F.~Piccoli, R.~Schettini, Artistic photo filter removal
  using convolutional neural networks, Journal of Electronic Imaging 27~(1)
  (2017) 011004.

\bibitem{bianco2019learning}
S.~Bianco, C.~Cusano, F.~Piccoli, R.~Schettini, Learning parametric functions
  for color image enhancement, in: International Workshop on Computational
  Color Imaging, Springer, 2019, pp. 209--220.

\bibitem{bianco2020personalized}
S.~Bianco, C.~Cusano, F.~Piccoli, R.~Schettini, Personalized image enhancement
  using neural spline color transforms, IEEE Transactions on Image Processing
  29 (2020) 6223--6236.

\bibitem{zhang2019zero}
L.~Zhang, L.~Zhang, X.~Liu, Y.~Shen, S.~Zhang, S.~Zhao, Zero-shot restoration
  of back-lit images using deep internal learning, in: Proceedings of the 27th
  {ACM} International Conference on Multimedia, 2019, pp. 1623--1631.

\bibitem{chai2020supervised}
Y.~Chai, R.~Giryes, L.~Wolf, Supervised and unsupervised learning of
  parameterized color enhancement, in: Proceedings of the IEEE/CVF Winter
  Conference on Applications of Computer Vision, 2020, pp. 992--1000.

\bibitem{kim2021representative}
H.~Kim, S.-M. Choi, C.-S. Kim, Y.~J. Koh, Representative color transform for
  image enhancement, in: Proceedings of the IEEE/CVF International Conference
  on Computer Vision, 2021, pp. 4459--4468.

\bibitem{guo2020zero}
C.~Guo, C.~Li, J.~Guo, C.~C. Loy, J.~Hou, S.~Kwong, R.~Cong, Zero-reference
  deep curve estimation for low-light image enhancement, in: Proceedings of the
  IEEE/CVF Conference on Computer Vision and Pattern Recognition, 2020, pp.
  1780--1789.

\bibitem{park2018distort}
J.~Park, J.-Y. Lee, D.~Yoo, I.~S. Kweon, Distort-and-recover: Color enhancement
  using deep reinforcement learning, in: Proceedings of the {IEEE} conference
  on computer vision and pattern recognition, 2018, pp. 5928--5936.

\bibitem{yang2018personalized}
H.~Yang, B.~Wang, N.~Vesdapunt, M.~Guo, S.~B. Kang, Personalized exposure
  control using adaptive metering and reinforcement learning, IEEE transactions
  on visualization and computer graphics 25~(10) (2018) 2953--2968.

\bibitem{hu2018exposure}
Y.~Hu, H.~He, C.~Xu, B.~Wang, S.~Lin, Exposure: A white-box photo
  post-processing framework, ACM Transactions on Graphics (TOG) 37~(2) (2018)
  1--17.

\bibitem{yu2018crafting}
K.~Yu, C.~Dong, L.~Lin, C.~C. Loy, Crafting a toolchain for image restoration
  by deep reinforcement learning, in: Proceedings of the {IEEE} conference on
  computer vision and pattern recognition, 2018, pp. 2443--2452.

\bibitem{yu2018deepexposure}
R.~Yu, W.~Liu, Y.~Zhang, Z.~Qu, D.~Zhao, B.~Zhang, Deepexposure: Learning to
  expose photos with asynchronously reinforced adversarial learning, in:
  Proceedings of the 32nd International Conference on Neural Information
  Processing Systems, 2018, pp. 2153--2163.

\bibitem{cai2022mask}
Y.~Cai, J.~Lin, X.~Hu, H.~Wang, X.~Yuan, Y.~Zhang, R.~Timofte, L.~Van~Gool,
  Mask-guided spectral-wise transformer for efficient hyperspectral image
  reconstruction, in: Proceedings of the IEEE/CVF Conference on Computer Vision
  and Pattern Recognition, 2022, pp. 17502--17511.

\bibitem{lin2022coarse}
J.~Lin, Y.~Cai, X.~Hu, H.~Wang, X.~Yuan, Y.~Zhang, R.~Timofte, L.~Van~Gool,
  Coarse-to-fine sparse transformer for hyperspectral image reconstruction,
  arXiv preprint arXiv:2203.04845 (2022).

\bibitem{zhang2021star}
Z.~Zhang, Y.~Jiang, J.~Jiang, X.~Wang, P.~Luo, J.~Gu, Star: A structure-aware
  lightweight transformer for real-time image enhancement, in: Proceedings of
  the IEEE/CVF International Conference on Computer Vision, 2021, pp.
  4106--4115.

\bibitem{browne2012survey}
C.~B. Browne, E.~Powley, D.~Whitehouse, S.~M. Lucas, P.~I. Cowling,
  P.~Rohlfshagen, S.~Tavener, D.~Perez, S.~Samothrakis, S.~Colton, A survey of
  monte carlo tree search methods, IEEE Transactions on Computational
  Intelligence and AI in games 4~(1) (2012) 1--43.

\bibitem{silver2018masteringzero}
D.~Silver, T.~Hubert, J.~Schrittwieser, I.~Antonoglou, M.~Lai, A.~Guez,
  M.~Lanctot, L.~Sifre, D.~Kumaran, T.~Graepel, T.~Lillicrap, K.~Simonyan,
  D.~Hassabis, A general reinforcement learning algorithm that masters {Chess},
  {Shogi}, and {Go} through self-play, Science 362~(6419) (2018) 1140--1144.

\bibitem{Chen2018Retinex}
W.~Chen, W.~Wenjing, Y.~Wenhan, L.~Jiaying, Deep retinex decomposition for
  low-light enhancement, in: British Machine Vision Conference, 2018.

\bibitem{lim2020dslr}
S.~Lim, W.~Kim, Dslr: Deep stacked laplacian restorer for low-light image
  enhancement, IEEE Transactions on Multimedia (2020).

\bibitem{LIU2022109039}
X.~Liu, W.~Ma, X.~Ma, J.~Wang, Lae-net: A locally-adaptive embedding network
  for low-light image enhancement, Pattern Recognition (2022) 109039.

\bibitem{gharbi2017deep}
M.~Gharbi, J.~Chen, J.~T. Barron, S.~W. Hasinoff, F.~Durand, Deep bilateral
  learning for real-time image enhancement, ACM Transactions on Graphics (TOG)
  36~(4) (2017) 1--12.

\bibitem{fivek}
V.~Bychkovsky, S.~Paris, E.~Chan, F.~Durand, Learning photographic global tonal
  adjustment with a database of input / output image pairs, in: The
  Twenty-Fourth IEEE Conference on Computer Vision and Pattern Recognition,
  2011.

\bibitem{zhang2018perceptual}
R.~Zhang, P.~Isola, A.~A. Efros, E.~Shechtman, O.~Wang, The unreasonable
  effectiveness of deep features as a perceptual metric, in: CVPR, 2018.

\end{thebibliography}

\end{document}